\documentclass[11pt]{article}

\usepackage[T1]{fontenc}
\usepackage[utf8]{inputenc}
\usepackage{microtype}
\usepackage{geometry}
\usepackage{amsmath, amssymb}
\usepackage{booktabs}
\usepackage{graphicx}
\usepackage{url}
\usepackage{array}
\usepackage{enumitem}
\usepackage{float}
\usepackage{hyperref}

\title{Real-Time In-Cabin Driver Behavior Recognition on Low-Cost Edge Hardware}

\author{
Vesal Ahsani, Babak H.~Khalaj, and Hamed Shah-Mansouri\\
Department of Electrical Engineering, Sharif University of Technology, Tehran, Iran\\
\texttt{vesal.ahsani@ee.sharif.edu, khalaj@sharif.edu, hamedsh@sharif.edu}
}

\date{} 

\begin{document}
\maketitle

\begin{abstract}
In-cabin driver monitoring systems (DMS) must recognize distraction- and drowsiness-related behaviors with low latency under strict constraints on compute, power, and cost. We present a single-camera in-cabin driver behavior recognition system designed for deployment on two edge platforms: Raspberry Pi~5 (CPU-only) and the Google Coral development board with an Edge Tensor Processing Unit (Edge~TPU) accelerator. The proposed pipeline combines (i) a compact per-frame vision model, (ii) a confounder-aware label design to reduce confusions among visually similar behaviors, and (iii) a temporal decision head that triggers alerts only when predictions are both confident and sustained. The system covers 17 distinct behavior classes. Training and evaluation use licensed datasets plus in-house collection (total $>800{,}000$ labeled frames) with driver-disjoint splits, and we further validate the deployed system in live in-vehicle tests. End-to-end performance reaches $\sim$16 frames per second (FPS) on Raspberry Pi~5 using 8-bit integer (INT8) inference (per-frame latency $<60$\,ms) and $\sim$25 FPS on Coral Edge~TPU (end-to-end latency $\sim$40\,ms), enabling real-time monitoring and stable alert generation on embedded hardware. Finally, we discuss how reliable in-cabin perception can serve as an upstream signal for human-centered vehicle intelligence, including agentic vehicle concepts~\cite{yu2025agv}.
\end{abstract}


\section{Introduction}
Driver Monitoring Systems (DMS) are becoming a core safety component in modern vehicles, enabling distraction
detection and drowsiness alerting and increasingly serving as a prerequisite for advanced driver-assistance
systems (ADAS) and higher levels of vehicle automation. Drowsiness and distraction remain persistent contributors
to crash risk: population-scale surveys report that roughly \emph{1 in 25} adults have fallen asleep while driving
in the preceding month \cite{cdc_drowsy_2013}, and U.S. summaries attribute hundreds of annual fatalities to
drowsy-driving involvement \cite{nhtsa_drowsy_2021}. Similarly, national reporting continues to document thousands
of fatalities in distraction-affected crashes \cite{nhtsa_distracted_2021}. These safety stakes have accelerated
interest in robust, always-on in-cabin perception that can operate continuously and reliably under real driving
conditions, where risk-relevant behaviors may be subtle, short-lived, partially occluded, or visually ambiguous.

At the same time, the operational context of modern ADAS increases the importance of in-cabin monitoring. Partial
driving automation (e.g., SAE Level~2) can control both lateral and longitudinal motion under specific conditions,
but the human driver remains responsible for supervision and timely intervention \cite{sae_j3016}. In practice,
driver understanding of assistance-system limitations can be incomplete; for example, surveys of Adaptive Cruise
Control (ACC) owners report substantial unawareness of key limitations \cite{aaa_adas_report}. This mismatch
between system capability and driver mental models motivates monitoring approaches that can estimate attention
allocation and secondary-task engagement in real time. Regulatory and consumer-assessment trends further reinforce
this direction: the EU General Safety Regulation framework includes driver drowsiness and attention warning (DDAW)
requirements \cite{eu_2019_2144}, and safety assessment protocols explicitly incentivize effective monitoring that
supports safe driving behavior while controlling false-positive rates \cite{euroncap_safe_driving}.

Despite rapid progress in vision-based recognition, real-world in-cabin monitoring remains challenging.
Illumination can shift abruptly (e.g., tunnels, shadows, sunset glare, night driving, oncoming headlights),
drivers vary substantially in appearance and habits, and hands and faces are frequently occluded by steering wheels,
phones, beverages, sunglasses, or self-occluding poses. Real vehicles also introduce system-level disturbances that
are easy to underestimate in offline experiments: vibration-induced motion blur, rolling-shutter artifacts, and
small viewpoint changes caused by mounting geometry. These factors are amplified on embedded platforms, where limited
compute budgets restrict model capacity, input resolution, and temporal modeling complexity. As a result, a recurring
gap persists between academic prototypes and deployable DMS systems.

A key reason for this gap is that many studies emphasize offline frame-level accuracy under controlled or
benchmark-centric settings, while an operational DMS must behave as an end-to-end system that:
(i) runs in real time on low-cost edge hardware,
(ii) limits false alarms caused by visually similar actions (confounders), and
(iii) produces stable alert events suitable for human feedback and downstream integration, rather than noisy
frame-by-frame labels. In practice, even a high-accuracy classifier can be unusable if it generates frequent
nuisance alerts or oscillates between confusable classes during natural transient motions (brief glances, short
gestures, momentary hand-to-face movement). This motivates designs that treat deployment as a system problem and
explicitly bridge the gap between per-frame recognition and event-level alerting.

This paper presents an end-to-end single-camera in-cabin behavior recognition system designed explicitly around
these operational requirements and validated in real in-vehicle tests on two low-cost edge platforms. Our pipeline
combines a compact per-frame vision model with two system-level mechanisms aimed at deployment robustness:
(1) a confounder-aware behavior taxonomy that explicitly models common look-alike activities that otherwise trigger
false positives, and (2) a temporal decision head that converts frame-level predictions into event-level alerts
using confidence and persistence gating. The result is a system that prioritizes stable, actionable alerts under real
disturbances while maintaining near real-time performance on inexpensive hardware.

\paragraph{Why in-cabin sensing can matter beyond alerts.}
Recent work on \emph{agentic vehicles} (AgVs) argues that future vehicle intelligence may extend beyond task
automation toward goal-aware, human-centered interaction and adaptation \cite{yu2025agv}. Independent of whether
such agency is realized through explicit agentic architectures, a practical requirement for human-centered
decision-making is reliable estimation of occupant state and behavior under real constraints. In this sense,
in-cabin behavior recognition can serve as an upstream signal for higher-level decision layers (e.g., detecting
driver readiness to take over, identifying sustained secondary-task engagement, or distinguishing transient glances
from prolonged distraction), while still remaining useful as a standalone safety function. This positioning motivates
an emphasis on robustness mechanisms that reduce confusions and stabilize alerts \emph{before} such signals are
consumed by downstream logic.

\subsection{Contributions}
This work makes the following contributions:
\begin{itemize}[leftmargin=*]
  \item \textbf{An end-to-end deployable DMS pipeline designed for robustness, not only offline accuracy.}
  We present a single-camera system that couples per-frame recognition with a temporal decision head, producing stable
  event-level alerts under real driving disturbances (illumination shifts, vibration, occlusion, transient gestures).

  \item \textbf{A confounder-aware 17-behavior taxonomy for practical in-cabin monitoring.}
  The label design includes high-impact distraction/drowsiness behaviors as well as frequent confounders observed in real driving,
  reducing visually induced false positives during deployment.

  \item \textbf{Edge-optimized real-time deployment on low-cost hardware.}
  We report end-to-end in-vehicle performance on Raspberry Pi~5 (CPU-only, INT8, $\sim$16 FPS, per-frame latency below 60 milliseconds)
  and the Coral development board with Edge~TPU ($\sim$25 FPS), demonstrating real-time monitoring and stable alert generation on
  inexpensive embedded platforms.
\end{itemize}

\section{Problem Setup and Design Goals}
\label{sec:setup}
We study real-time in-cabin driver behavior recognition using a \emph{single} forward-facing monocular camera mounted to observe the driver during natural driving. Depending on the deployment setting, the sensor operates either in visible-spectrum
(RGB) mode or near-infrared (NIR) mode with active IR illumination for night robustness. The target use case is continuous on-vehicle operation, where the
system must remain functional under the practical disturbances that dominate real deployments: vehicle motion
(vibration and brief blur), rapid illumination transitions (shadows/tunnels, sunset glare, night scenes),
partial occlusions (hands, steering wheel, phone/cup), and moderate viewpoint variation due to mounting geometry.
These factors are also explicitly reflected in contemporary assessment and approval frameworks for driver
monitoring / driver engagement, which emphasize robustness across drivers and conditions and include occlusion
and accessories testing \cite{euroncap_driverengagement,euroncap_sd202}. Related regulatory requirements for
driver drowsiness and attention warning (DDAW) systems further motivate an end-to-end, system-level evaluation
beyond inference-only benchmarks \cite{eu_2021_1341,unece_ddaw}.

\paragraph{Streaming formulation.}
Let $x_t \in \mathbb{R}^{H \times W \times 3}$ denote the RGB frame at discrete time index $t$, sampled with
period $\Delta$ (seconds). Let $y_t \in \{1,\dots,C\}$ be the frame-level ground-truth behavior label, where
$C=17$ corresponds to the behavior set in Table~\ref{tab:behaviors}. The perception module produces a probability
simplex vector
\begin{equation}
\mathbf{p}_t = f_\theta(x_t), \qquad p_t(c)=\Pr(y_t=c \mid x_t), \quad c\in\{1,\dots,C\},
\end{equation}
where $f_\theta$ is a compact vision model chosen to satisfy embedded execution constraints (Section~\ref{sec:deployment}).
Although the backbone is per-frame, the deployed system operates on a stream and may exploit short temporal context
through buffering and decision logic downstream (Section~\ref{sec:temporalhead}).

\paragraph{From recognition to operational alerting.}
A deployable DMS requires an interface that is stable enough for driver feedback and downstream integration.
Directly emitting $\hat{y}_t=\arg\max_c p_t(c)$ is brittle in real driving: transient motions and brief occlusions
can create isolated probability spikes that fragment alerts and increase nuisance warning rates. In practice, attention-warning
algorithms and assessment protocols explicitly distinguish between short driving-related glances and sustained off-road engagement,
highlighting that the relevant output is temporal and policy-dependent rather than frame-by-frame labels \cite{forster2024_attentionwarn,euroncap_driverengagement}.

We therefore define an alert output as a set of \emph{events}
\begin{equation}
\mathcal{A}=\{(c, t_\text{start}, t_\text{end})\},
\end{equation}
where each tuple denotes a detected instance of behavior $c$ over an interval $[t_\text{start}, t_\text{end}]$.
Event formation is performed by a temporal decision head that enforces confidence and persistence constraints
(Section~\ref{sec:temporalhead}). This design treats alerting as a sequential decision problem with an explicit
trade-off between sensitivity (fast detection) and specificity (low nuisance alarms).

\paragraph{End-to-end timing model and jitter.}
Because a streaming DMS is a closed-loop real-time system, we treat deployment as an end-to-end timing problem.
Let the per-frame end-to-end latency be
\begin{equation}
L^{\mathrm{e2e}}_t = L_t^{\text{cap}} + L_t^{\text{pre}} + L_t^{\text{inf}} + L_t^{\text{post}} + L_t^{\text{io}},
\end{equation}
corresponding to capture/decode, preprocessing, inference, postprocessing (including temporal logic), and overlay/event emission.
We denote sustained throughput as $\text{FPS} \approx 1/\text{median}(L^{\mathrm{e2e}}_t)$ and quantify timing stability via tail latency
(e.g., $p95(L^{\mathrm{e2e}}_t)$) to capture jitter that can destabilize persistence-based decision logic.
This system-level view is aligned with evaluation practices in driver monitoring protocols that require evidence of stable operation
under a range of circumstances, including degraded sensing conditions \cite{euroncap_driverengagement,euroncap_sd202}.

\paragraph{Design goals.}
The system is designed for deployment rather than offline benchmarking, with four concrete goals:
\begin{itemize}[leftmargin=*]
  \item \textbf{Real-time operation on low-cost edge hardware (system-level).}
  Sustain throughput sufficient for temporal persistence (targeting $\sim$15--25 FPS) with bounded end-to-end latency
  (tens of milliseconds per frame), accounting for camera I/O and application logic (not inference-only timing).

  \item \textbf{Robustness to in-cabin distribution shifts.}
  Maintain usable performance across lighting regimes, driver appearance/accessories, partial occlusions, and moderate viewpoint
  variation. From a safety standpoint, these conditions correspond to \emph{reasonably foreseeable} operating scenarios and
  misuse patterns that must be handled without introducing unreasonable risk due to functional insufficiency \cite{iso_21448}.

  \item \textbf{Reduced confusions among visually similar actions (confounders).}
  Explicitly model frequent look-alike behaviors (e.g., grooming vs.\ phone use; control-panel interaction vs.\ texting) to reduce
  systematic false positives that can dominate nuisance warning rates in continuous operation.

  \item \textbf{Stable event-level alerting with tunable sensitivity.}
  Produce a temporally consistent alert stream whose operating point (confidence threshold, persistence window, cooldown) can be
  tuned to meet application constraints and evaluation criteria that penalize excessive false warnings \cite{euroncap_driverengagement,forster2024_attentionwarn}.
\end{itemize}

\paragraph{Key deployment constraints.}
We target low-cost embedded hardware under compute and thermal limits. Practically, this implies:
(i) sustained throughput to support persistence windows on the order of $\sim$0.5--1.0 seconds,
(ii) low end-to-end latency for timely warnings,
(iii) resilience to brief occlusions and transient motions without spurious triggers, and
(iv) predictable timing (low jitter) to avoid instability in event formation under streaming operation.

\section{Related Work}
\label{sec:rw}

Driver monitoring has evolved from early cue-specific pipelines (e.g., eye closure and gaze heuristics) to
learning-based in-cabin perception models that recognize distraction and drowsiness from images or short video
windows. At the same time, practical deployment on low-cost embedded hardware has made end-to-end system design
(latency, jitter, operator compatibility, nuisance alerts) as important as offline recognition accuracy. We review
relevant work across four themes: (i) classical cue-based DMS, (ii) datasets/benchmarks for in-cabin monitoring,
(iii) learning-based behavior recognition and temporal modeling, and (iv) efficient edge deployment and alert stability.

\subsection{Cue-based driver monitoring: gaze, eye closure, and landmark heuristics}
Early and still-common DMS designs estimate driver vigilance using interpretable physiological/kinematic cues such as
head pose, gaze direction, blink dynamics, and eyelid closure measures. A widely used indicator of drowsiness is
PERCLOS (percentage of eyelid closure over time), which has been studied as a physiologically meaningful proxy for
reduced alertness and fatigue-related performance degradation \cite{dinges1998perclos}. These pipelines often rely on
facial landmarks or eye/mouth geometry and apply heuristic thresholding and temporal rules.

With the availability of robust real-time landmark tracking toolkits, landmark-driven methods remain attractive for
embedded deployments due to their low compute overhead. MediaPipe, for example, provides a practical framework for
building real-time perception pipelines with on-device face/landmark tracking \cite{lugaresi2019mediapipe}. For blink
and eye-state estimation, the Eye Aspect Ratio (EAR) formulation is frequently used as a lightweight geometric
signal; Soukupov\'a and \v{C}ech provide a widely cited real-time approach for blink detection from facial landmarks
under standard camera input \cite{soukupova2016ear}. However, landmark-only approaches can degrade under precisely the
conditions that dominate in real in-cabin monitoring: sunglasses, masks/face coverings, hands occluding the face,
strong illumination changes, and motion blur. These limitations motivate complementary appearance-based recognition
models that can exploit broader context beyond sparse landmarks.

\subsection{Datasets and benchmarks for in-cabin monitoring}
Progress in learning-based DMS has been accelerated by public datasets covering distraction and drowsiness under
controlled and naturalistic conditions. For fine-grained in-cabin activity recognition, Drive\&Act provides a large
multi-modal and multi-view benchmark with a hierarchical label space and synchronized RGB/IR/depth/pose streams
collected under both manual and automated driving \cite{martin2019driveact}. In the drowsiness domain, datasets such
as NTHU-DDD include day/night recordings of drivers with staged drowsiness cues and have been widely used for
benchmarking video-based drowsiness detection \cite{nthu_ddd}. For yawning analysis specifically, YawDD provides
driver-facing video sequences collected under varying illumination and facial characteristics and has supported
comparative evaluation of yawning-related fatigue cues \cite{abtahi2014yawdd}.

For distraction/posture recognition, several datasets emphasize real-world variability in drivers, vehicles, and camera
views. Abouelnaga et al.\ introduced a practical distracted-driving posture classification setting that has been widely
used in follow-on work \cite{abouelnaga2017distracted}. More recently, 100-Driver substantially scales driver diversity
and viewpoint variation, providing multi-camera coverage and a large number of images across drivers/vehicles to
stress-test generalization under realistic shifts \cite{wang2023hundreddriver}. DMD further expands coverage with a
multi-modal dataset designed for attention and alertness analysis and includes a derived behavior subset (dBehaviourMD)
for distraction activities, explicitly highlighting the role of multi-camera and multi-sensor signals in robust DMS
development \cite{ortega2020dmd}.

These benchmarks illustrate an important tension for deployable systems: the richest datasets often rely on multiple
views or modalities, while cost-constrained deployments frequently require single-camera solutions with predictable
latency and minimal calibration. Our work targets this single-camera operating point while emphasizing robustness
mechanisms (confounder-aware labeling and event-level alert gating) that are crucial under real driving disturbances.

\subsection{Learning-based in-cabin behavior recognition and temporal modeling}
Modern DMS increasingly treats distraction and drowsiness recognition as a supervised learning problem on images or
short video windows. For drowsiness detection, temporal modeling has been explored via long-term or multi-granularity
architectures that incorporate short-term cues (blink/yawn) and longer context \cite{lyu2018longterm}. Condition-adaptive
representation learning has also been proposed to improve robustness across changing conditions and drivers
\cite{yu2019conditionadaptive}, and two-stream/multi-feature designs highlight the benefit of combining complementary
visual cues for fatigue-related behaviors \cite{shen2020drowsy}. For distraction and in-cabin actions, multi-modal
recognition benchmarks such as Drive\&Act motivate hierarchical behavior representations and fine-grained action
categories \cite{martin2019driveact}.

A practical deployment issue is that per-frame predictions can fluctuate rapidly due to vibration, illumination flicker,
and transient gestures. Many systems use some form of temporal aggregation (e.g., smoothing, majority vote, hidden-state
rules), but the alert interface itself is often under-specified. This matters because operational DMS performance is
determined not only by classification accuracy but also by nuisance-alert rate, fragmentation, and time-to-detect.
Motivated by human-factors findings that acceptance depends strongly on perceived reliability and nuisance alarms
\cite{donmez2006attitudes}, our work explicitly separates per-frame recognition from alert generation via a persistence-
and-confidence decision head (Section~\ref{sec:temporalhead}), producing event-level alerts that are more integration-ready.

\subsection{Efficient edge deployment and end-to-end system constraints}
Edge deployment imposes constraints that are frequently glossed over in offline evaluation: operator support, quantization
behavior, memory bandwidth, camera I/O, and timing jitter. Mobile-oriented architectures such as MobileNetV3 and efficiency-
driven scaling approaches such as EfficientNet provide strong accuracy--latency trade-offs for on-device vision
\cite{howard2019mobilenetv3,tan2019efficientnet}. Post-training integer quantization is a widely used tool for reducing
compute and memory footprint on embedded platforms \cite{jacob2018quantization}, and TensorFlow Lite is a common deployment
runtime for mobile and embedded inference \cite{tflite}. For accelerator-based deployment, Edge-TPU execution can provide
stable low latency when the graph is fully compatible with compiler/operator constraints \cite{coralbenchmarks}.

Several recent systems demonstrate different points in the design space. Khalil et al.\ report a low-cost DMS on an
ultra-cheap Raspberry Pi platform with coarse driver-state categories \cite{khalil2025lowcost}. Hariharan et al.\ report
high-throughput real-time driver monitoring on an automotive-oriented accelerator platform and discuss deployment issues
including operator constraints and model modifications \cite{hariharan2023edgedms}. Ortega et al.\ also emphasize real-time
considerations in multi-modal DMS pipelines and provide baselines that aim to be compatible with cost-efficient CPU-only
settings \cite{ortega2020dmd}. In contrast, our work focuses on a single-camera, 17-class behavior set and explicitly
treats deployment as an end-to-end system problem, reporting live in-vehicle measurements (capture-to-alert) and using
confounder-aware labeling plus persistence-gated alerting to improve operational stability on low-cost hardware.

\section{System Overview}
\label{sec:system}

\subsection{Target hardware}
We evaluate two low-cost deployment targets that are representative of practical edge constraints in vehicle pilots:
(i) a CPU-only embedded platform, and (ii) an embedded platform equipped with a dedicated neural inference accelerator.
Using these two targets allows us to quantify the performance of the same end-to-end pipeline under two realistic
compute regimes and to report deployment behavior that is directly relevant to cost-constrained prototyping and
early-stage product validation.

\begin{itemize}[leftmargin=*]
  \item \textbf{Raspberry Pi~5 (CPU-only):} a low-cost general-purpose embedded computer that supports flexible integration
  with camera devices and development tooling on embedded Linux. This target represents the common baseline scenario
  where inference must run on the CPU under tight thermal and power limits.

  \item \textbf{Google Coral Dev Board (Edge-TPU):} an embedded platform with an Edge-TPU accelerator intended for
  low-latency inference with supported TensorFlow Lite graphs and operators. This target represents an accelerator-enabled
  deployment track where throughput and timing stability can improve substantially when the model is fully mapped to the accelerator
  \cite{coralbenchmarks}.
\end{itemize}

For both platforms, we measure end-to-end throughput (FPS) and per-frame latency during real in-vehicle operation. For on-road evaluation, we use a single NIR in-cabin camera with integrated IR LEDs to enable stable imaging at night and in low light. End-to-end measurement includes camera capture, preprocessing, inference, postprocessing (including temporal alert logic),
and visualization/event emission, rather than reporting inference-only timing.

\subsection{End-to-end pipeline}
\begin{figure}[H]
  \centering
  \includegraphics[width=\linewidth]{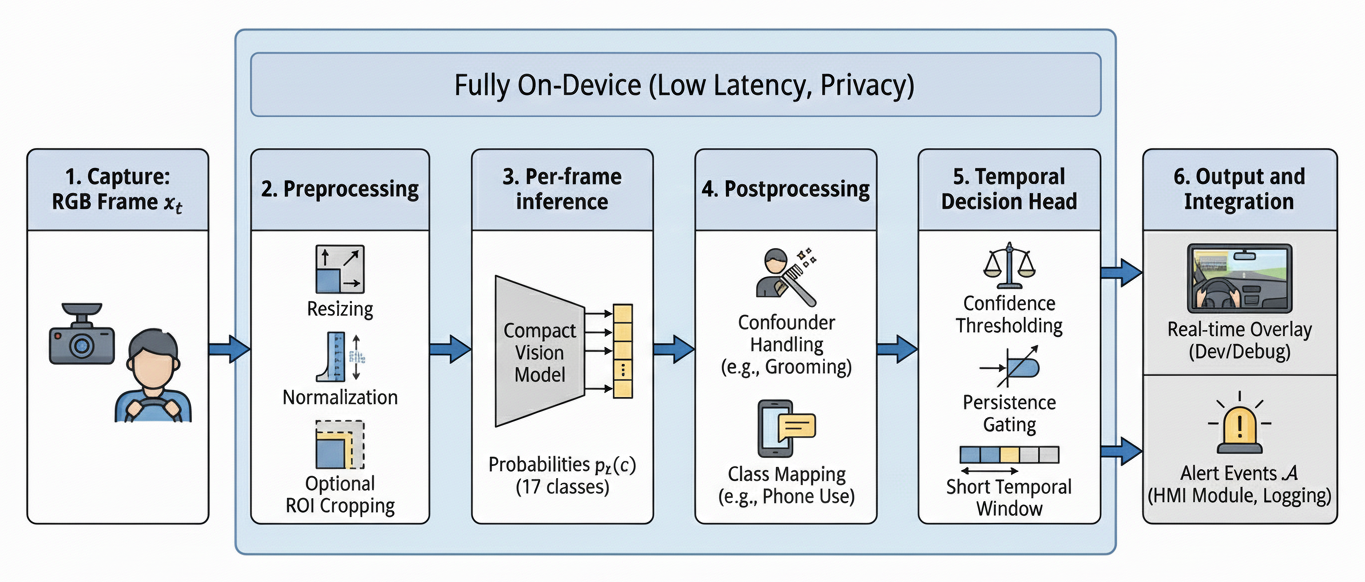}
  \caption{End-to-end in-cabin driver behavior recognition pipeline (conceptual).}
  \label{fig:pipeline}
\end{figure}

Figure~\ref{fig:pipeline} summarizes the processing pipeline that converts a stream of RGB frames into event-level behavior alerts.
The pipeline is designed to remain fully on-device to ensure low latency and predictable timing, and to avoid reliance on cloud
round-trips that would introduce variable delays and additional privacy exposure.

\begin{enumerate}[leftmargin=*]
  \item \textbf{Capture.} Acquire an RGB frame $x_t$ from a camera facing the driver at time index $t$.
  The camera is mounted in a dashboard / rear-view-mirror region and provides a forward-facing view of the driver.

  \item \textbf{Preprocessing.} Apply lightweight preprocessing suitable for embedded operation, including resizing to the
  model input resolution and normalization consistent with the training pipeline. When applicable, optional region-of-interest
  strategies can be used to reduce irrelevant background content and improve stability (e.g., cropping to the driver area),
  while keeping the pipeline compatible with real-time constraints.

  \item \textbf{Per-frame inference.} Apply a compact vision model to compute per-frame class probabilities $p_t(c)$ over the
  17 behavior classes. The per-frame model provides the primary recognition signal, while temporal logic is handled downstream
  to improve operational stability.

  \item \textbf{Postprocessing and confounder handling.} Convert model outputs into application-level behavior signals.
  This step leverages explicit confounder classes included in the taxonomy (e.g., grooming, control-panel interaction) and can
  optionally apply class mapping to match a target alert policy (for example, collapsing multiple phone-use modes into a single
  phone-use alert if desired by an application).

  \item \textbf{Temporal decision head.} Convert frame-level probabilities into event-level alerts by applying confidence
  thresholding and persistence gating over a short temporal window (Section~\ref{sec:temporalhead}). This step is critical
  for suppressing spurious triggers caused by transient gestures, brief glances, and momentary occlusions.

  \item \textbf{Output and integration.} Render a real-time overlay for development and debugging, and emit alert events
  $\mathcal{A}$ for downstream integration (e.g., logging, warning logic, or a human--machine interface (HMI) module).
\end{enumerate}

\section{Behavior Set (17 Classes)}
\label{sec:behaviors}
We formulate in-cabin behavior recognition as a single-label classification problem over a practical set of driver actions and
states that commonly affect safety and attention. While real driving can involve overlapping activities, a single-label formulation
is often appropriate for real-time embedded monitoring because it enables stable alert policies that prioritize the most salient
safety-relevant behavior at each time step.

The behavior taxonomy is designed with two objectives. First, it covers high-impact distraction and drowsiness cues that are meaningful
for alerting. Second, it explicitly includes frequent confounding behaviors observed in real driving (e.g., grooming/hand-on-hair or
control-panel interaction) that can visually resemble unsafe behaviors and otherwise inflate nuisance alarms if not modeled as separate
classes.

In deployment, the per-frame classifier outputs a label (or probability distribution) at each frame, while alerting is produced at the
event level via temporal persistence (Section~\ref{sec:temporalhead}). This separation allows the taxonomy to be fine-grained for recognition,
while enabling downstream applications to define alert policies through mapping and gating (e.g., differentiating texting versus talking, or
collapsing both into a single phone-use alert depending on requirements).

The 17 classes in Table~\ref{tab:behaviors} can be grouped into: (a) phone interaction modes (talking vs.\ texting, left vs.\ right),
(b) consumption-related actions (eating, drinking, smoking), (c) gaze/attention shifts and interaction (look left, look down, talking to
passengers/look right), (d) cabin actions (reaching behind), and (e) drowsiness-related cues (yawning, eyes-closed sleep).
Left/right splits are included where relevant because viewpoint geometry and hand usage differ; in practice, separating these cases can reduce
confusion and improve stability during edge deployment.

We merge ``Talking to Passengers'' with ``Look Right'' because, under our camera geometry and annotation practice, passenger interaction is
primarily manifested as sustained head/gaze orientation toward the right-side passenger. Distinguishing gaze-only behavior from interaction-only
behavior is not reliably achievable without additional modalities (e.g., audio) or multi-view sensing.

\begin{table}[H]
\centering
\caption{Supported 17 behavior classes used for training and deployment.}
\label{tab:behaviors}
\begin{tabular}{cl}
\toprule
\textbf{ID} & \textbf{Behavior} \\
\midrule
1  & Normal (Safe driving) \\
2  & Phone -- Talk Left \\
3  & Phone -- Talk Right \\
4  & Phone -- Text Left \\
5  & Phone -- Text Right \\
6  & Eating \\
7  & Drinking \\
8  & Smoking Right \\
9  & Smoking Left \\
10 & Reaching Behind \\
11 & Look Left \\
12 & Look Down \\
13 & Talking to Passengers / Look Right \\
14 & Makeup / Hand on Hair \\
15 & Control Panel / GPS \\
16 & Yawning \\
17 & Sleep / Eyes Closed \\
\bottomrule
\end{tabular}
\end{table}

\section{Dataset and Diversity}
\label{sec:dataset}
Training and evaluation use a large-scale in-cabin dataset comprising more than 800{,}000 labeled images,
assembled from two licensed/commercial sources and additional in-house collection conducted by the authors.
The dataset covers the 17 behavior classes in Table~\ref{tab:behaviors} and is curated to reflect practical in-cabin
conditions encountered during real driving (illumination transitions, occlusions, and viewpoint variation).
Because parts of the dataset are proprietary, we cannot redistribute the raw images. To support transparency and
comparability, we report dataset scale, the major sources of variation, and the split protocol used to reduce leakage
and better approximate real-world generalization.

\subsection{Scale, labeling, and split protocol}
All samples are annotated at the frame level for the 17 behavior classes listed in Table~\ref{tab:behaviors}.
The labeled frames include both short transient behaviors (e.g., brief glance shifts) and sustained actions
(e.g., phone interaction), which is important for evaluating the temporal alerting mechanism described in
Section~\ref{sec:temporalhead}.

We use an 80/10/10 split by labeled frames for train/validation/test, while enforcing a \emph{driver-disjoint} partitioning
to prevent identity leakage across splits. The resulting partitions contain $150$ / $15$ / $20$ distinct drivers,
respectively, with no driver overlap. To further reduce leakage due to near-duplicate frames, we perform splitting at the
driver level (and, when available, at the vehicle/session/clip level) rather than at the individual-frame level.

\subsection{Diversity and realism factors}
The dataset is designed to increase diversity and robustness by spanning variation across drivers, vehicles, and conditions
that are known to affect in-cabin perception:

\paragraph{Driver diversity.}
The dataset includes male and female drivers, a range of ages, and diverse appearance factors such as clothing styles and
hairstyles. Common accessories and occlusions are represented, including glasses and sunglasses, hats, face masks, and other
face coverings. These factors are important because they can alter the visibility of facial cues, head pose, and eye region
appearance, and can increase ambiguity between visually similar behaviors.

\paragraph{Vehicle diversity.}
The dataset includes more than 10 vehicle types, spanning sedans, vans, trucks, crossovers, and other passenger vehicles.
Camera placement is chosen to be consistent with realistic deployment, with a dashboard / rear-view-mirror style viewpoint
facing the driver. Variation in cabin geometry and mounting position introduces changes in framing, scale, and occlusion
patterns that are representative of real vehicle pilots.

\paragraph{Environment and lighting diversity.}
The dataset covers a range of environments and lighting conditions, including bright daytime sunlight, sunset and evening,
night and low-light scenes, and driving on both city streets and highways. It also includes diverse reflections and shadow
patterns and rapid lighting transitions (e.g., entering/exiting shaded areas), which are common failure cases for in-cabin vision.

\subsection{Hard-example enrichment and label refinement}
During iterative development and road testing, we observed that small objects and subtle cues (e.g., phone, cigarette, cup)
can dominate recognition errors and can contribute disproportionately to spurious alerts in deployment. To improve robustness,
we performed targeted data enrichment by adding additional frames emphasizing these small-object cues, especially for
safety-relevant and visually confusable behaviors.

We also curated a special set of ``mirror gaze'' examples motivated by on-road observations: brief mirror checks can be confused
with ``look left'' / ``look right'' distraction classes if labels are not carefully defined. We therefore include examples that
distinguish (i) brief mirror glances that occur during normal driving from (ii) sustained off-road glances that are more consistent
with distraction. This refinement supports a cleaner separation between normal glance behavior and distraction-like gaze shifts and
improves the operational stability of event-level alerting.

\subsection{Real-world validation}
In addition to offline evaluation on the held-out test split, we validate the deployed system in real in-vehicle trials using a live
camera stream on both target platforms. This step is important because performance can shift under real driving due to vibration,
illumination changes, rolling-shutter artifacts, and mounting differences that are not fully captured in offline data. We therefore
report end-to-end runtime behavior and qualitative alert stability observed during on-road operation in Section~\ref{sec:inv_eval}.

\section{Method}
\label{sec:method}

\subsection{Confounder-aware recognition design}
A major failure mode in in-cabin monitoring is confusion between visually similar actions (confounders),
where distinct behaviors share overlapping visual evidence. Common examples include brief hand-to-face motion
being confused with phone use, and infotainment interaction being confused with texting. In deployment,
these confusions are costly: even infrequent frame-level errors can translate into nuisance alerts when a
system runs continuously, degrading usability and perceived reliability.

We address confounders through a confounder-aware design at two complementary levels:

\paragraph{Taxonomy design.}
Instead of forcing frequent confounding activities into nearby ``unsafe'' categories, we include explicit
classes that represent them (e.g., grooming/hand-on-hair and control-panel interaction).
This design turns many false positives into correctly recognized alternative explanations, improving precision
for safety-critical classes without requiring additional sensors.
The taxonomy therefore plays a dual role: it defines the recognition task and also embeds prior knowledge about
which behaviors are likely to be confused under realistic cabin viewpoints and occlusions.

\paragraph{Decision design.}
We decouple recognition from alerting. Per-frame classification provides a probabilistic estimate of behavior,
but alert generation is handled by a temporal decision head (Section~\ref{sec:temporalhead}) that enforces
confidence and persistence constraints. This reduces spurious triggers caused by transient gestures, brief
glances, and momentary occlusions. The separation between recognition and alerting also supports flexible
deployment policies, where applications can choose to map multiple fine-grained classes to a smaller set of
alert categories (e.g., collapsing multiple phone-use modes into a single ``phone use'' alert if desired).

\subsection{Per-frame model and training protocol}
\label{sec:training}
We formulate in-cabin behavior recognition as a $C$-class image classification problem with $C=17$ on individual
frames. Given an input frame $x_t$, the network outputs a probability vector
$\mathbf{p}_t \in [0,1]^C$, where $p_t(c)$ denotes the predicted probability of class $c$ at time $t$.

\paragraph{Model exploration and selection.}
We evaluated multiple edge-suitable vision stacks for per-frame behavior recognition, including:
(i) mobile-oriented convolutional neural network (CNN) backbones (MobileNetV3-style) \cite{howard2019mobilenetv3},
(ii) EfficientNet-family models \cite{tan2019efficientnet}, and (iii) compact hierarchical convolutional networks
with multi-scale feature aggregation and a task-specific classification head.

Rather than selecting a backbone solely by offline accuracy, we performed deployment-aware model selection under
joint constraints that reflect real embedded operation:
(a) robustness under in-cabin variability (occlusion, low-light, viewpoint shifts),
(b) temporal stability after event gating (Section~\ref{sec:temporalhead}), and
(c) hardware executability under integer quantization, including operator compatibility and sustained throughput
on both CPU-only and accelerator-enabled targets.
This process is motivated by an empirical deployment reality: architectures that perform well offline may not
translate cleanly to edge runtimes if they require unsupported operators, exhibit unstable timing, or degrade
disproportionately under quantization.

\paragraph{Input resolution trade-off.}
Input resolution directly affects both recognition quality (small objects such as phones, cups, and cigarettes)
and computational cost. We swept input resolutions from $224$ to $640$ and found that $320 \times 320$ provides a
favorable operating point: it preserves discriminative hand/face cues while meeting real-time constraints on the
target platforms. Higher resolutions provided diminishing returns relative to the additional latency, while lower
resolutions more frequently degraded small-object sensitivity and increased confusions among fine-grained behaviors.

\paragraph{Architecture configuration.}
The final configuration uses a compact hierarchical convolutional backbone with multi-stage downsampling and feature
mixing, followed by global aggregation and a task-specific $C$-way classification head ($C=17$). This structure is a
practical fit for in-cabin monitoring, where discriminative cues can be localized and small (e.g., hand posture and
near-face objects), but the system must operate under bounded compute.

The resulting model has approximately $15.8$ million parameters and approximately 41.9 gigafloating-point operations
(GFLOPs) at $320 \times 320$ input resolution, remaining feasible for real-time inference on the target embedded
platforms after integer quantization.
\footnote{The implementation is instantiated using the Ultralytics YOLOv8 classification family (medium capacity),
with the ImageNet head replaced by a 17-class head.}

\paragraph{Loss function and class imbalance handling.}
In-cabin monitoring datasets are inherently imbalanced: normal driving dominates, while safety-critical behaviors
(e.g., drowsiness and phone use) are comparatively rare. To reduce majority-class dominance and improve minority-class
recall, we adopt a focal loss with class-dependent weighting. Class weights are computed from inverse class frequencies
and capped to prevent excessively large gradients from rare classes.

For a labeled sample at time $t$, the loss is
\[
\mathcal{L}_t = -\alpha_{y_t} (1 - p_t(y_t))^{\gamma} \log p_t(y_t),
\]
where $\alpha_{y_t}$ is the capped class weight for the ground-truth label and $\gamma$ is the focusing parameter.
We set $\gamma=1.5$ in our experiments. This formulation encourages the model to focus on harder examples while maintaining
stable optimization on large-scale imbalanced data.

\paragraph{Training protocol.}
Models are trained with mini-batch optimization using the Adam optimizer with a fixed learning rate of $5\times10^{-4}$ and batch size $32$.
Training is run for $50$ epochs with validation-based model selection (best checkpoint on the validation split).
To preserve left/right semantic consistency (e.g., phone use on the left versus right), horizontal and vertical flips are disabled.
Training is performed on NVIDIA A100 GPUs. This configuration provides a stable baseline for subsequent deployment optimization and evaluation.

\subsection{Temporal decision head: confidence and persistence}
\label{sec:temporalhead}
Frame-level predictions are noisy under real driving due to vibration, illumination flicker, partial occlusions,
and short-duration actions. Triggering alerts directly from $\arg\max_c \mathbf{p}_t(c)$ can therefore yield fragmented
and repetitive warnings that are difficult to integrate downstream and may increase nuisance alarms. Motivated by the
need for operationally stable alert streams (and the broader human-factors concern that nuisance alerts reduce acceptance
of imperfect mitigation systems \cite{donmez2006attitudes}), we treat alerting as a separate temporal decision layer.

Let $p_t(c)$ denote the predicted probability for class $c$ at time $t$. Given a confidence threshold $\tau$ and a
persistence window length $K$ (in frames), an alert for class $c$ is triggered at time $t$ if the class remains both
dominant and confident over a contiguous window:
\begin{equation}
\label{eq:gate}
\Big(p_{t-i}(c) \ge \tau\Big)\ \land\ \Big(c = \arg\max_j p_{t-i}(j)\Big)\quad \forall i \in \{0,1,\dots,K-1\}.
\end{equation}

Once an alert for class $c$ becomes active, we maintain the event until evidence drops below a lower release threshold
$\tau_{\text{off}}<\tau$ for $M$ consecutive frames (we use $\tau_{\text{off}}=0.60$ and $M=3$), providing hysteresis and reducing boundary jitter.

In practice, $K$ is chosen to correspond to roughly one second of video (e.g., $K=25$ at $\sim$25 FPS), while $\tau$
controls sensitivity (default $\tau=0.75$). This mechanism converts per-frame probabilities into event-level alerts that
better reflect sustained behaviors rather than transient motions.

\paragraph{Event formation and termination.}
When the gating condition in Eq.~\ref{eq:gate} is satisfied, the system enters an active-alert state for class $c$ and
emits an event interval $[t_\text{start}, t_\text{end}]$. The event terminates when the persistence condition is violated
for a sufficient number of frames (equivalently, when evidence for class $c$ is no longer sustained). This yields a compact
alert representation that is more suitable for logging and downstream logic than raw frame-level labels.

\paragraph{Cooldown / hysteresis.}
After an alert fires, we enforce a short cooldown interval (configurable) to avoid repeated triggers for a single sustained
event. This reduces alert fragmentation and improves perceived stability during continuous behaviors.

\section{Edge Deployment}
\label{sec:deployment}
We target two low-cost edge platforms representative of real in-vehicle pilots: a CPU-only Raspberry Pi~5 and
the Google Coral development board with an Edge-TPU accelerator. For both targets, the deployment objective is
to minimize end-to-end latency while maintaining stable throughput sufficient for temporally consistent alerting
(Section~\ref{sec:temporalhead}). In this setting, deployment performance is not determined by inference speed alone:
camera capture, preprocessing, postprocessing, and visualization/event emission can contribute non-trivially to
latency and timing jitter. We therefore treat deployment as a system problem and report end-to-end measurements
under live in-vehicle operation.

\subsection{Raspberry Pi~5 (CPU-only)}
On Raspberry Pi~5, per-frame inference on the CPU is the primary runtime bottleneck. To meet real-time constraints
under tight compute and thermal limits, we deploy an INT8-quantized version of the per-frame classifier. Quantization
reduces both arithmetic cost and memory bandwidth pressure, which are critical on embedded CPUs where performance is
often limited by cache behavior and memory throughput.

\paragraph{Quantization procedure.}
We apply post-training quantization using a representative calibration set sampled across behavior classes and lighting
conditions. Calibration inputs follow the same preprocessing steps as runtime inference (resize and normalization) to
reduce quantization mismatch. After quantization, we verify that the deployed model preserves the input/output interface
expected by the online pipeline (capture $\rightarrow$ preprocess $\rightarrow$ inference $\rightarrow$ postprocess
$\rightarrow$ temporal gating), ensuring functional equivalence at the system level.

\paragraph{End-to-end runtime.}
In real in-vehicle tests, the optimized INT8 deployment achieves approximately $\sim$16 FPS with per-frame latency below
60 milliseconds, enabling near real-time alert generation with sufficient temporal resolution for persistence gating.
We emphasize that these measurements reflect end-to-end execution rather than inference-only timing.

\subsection{Coral Edge-TPU}
\label{sec:coral}
Deploying the classifier on Coral requires conversion to a TFLite graph compatible with the Edge-TPU compiler.
Accelerator-based execution introduces additional constraints beyond model size: the graph must satisfy operator support
and quantization requirements, and partial accelerator mapping can induce unpredictable latency due to CPU fallback.

\paragraph{Model export pipeline.}
We use a staged export pipeline:
(PyTorch $\rightarrow$ ONNX $\rightarrow$ TensorFlow SavedModel $\rightarrow$ TFLite $\rightarrow$ Edge-TPU compiled TFLite).
This sequence provides a practical path to (i) preserve functional equivalence across toolchains, (ii) enable post-training
integer quantization, and (iii) satisfy Edge-TPU operator constraints required for accelerator-only execution. In practice,
conversion stability and operator compatibility are as important as model size, and the staged pipeline makes these
constraints explicit and reproducible.

\paragraph{Post-training INT8 quantization.}
To minimize latency and maximize accelerator utilization, we apply full-integer quantization using a representative
calibration set (typically 200--500 images sampled across classes and lighting conditions). Calibration inputs follow
the same runtime preprocessing pipeline (resize and normalization) to reduce quantization mismatch and to preserve
performance after conversion.

\paragraph{Compilation and operator mapping.}
The quantized TFLite model is compiled using the Edge-TPU compiler. We validate the compilation report to confirm
accelerator-only execution (i.e., no CPU fallback). This step is critical because partial mapping can introduce timing
jitter and reduce effective throughput, which directly degrades the stability of persistence-based alerting. In our final
deployment graph, the compiler reports that all operations are mapped to the Edge-TPU. Table~\ref{tab:edgetpu_map}
summarizes the operator-level mapping for the final compiled model. In real in-vehicle tests, the compiled model sustains
approximately $\sim$25 FPS on the Coral development board, supporting low-latency temporal alerting. 

\begin{table}[h]
\centering
\caption{Edge-TPU compiler mapping summary for the final compiled model. All listed operators are mapped to the Edge-TPU (no CPU fallback).}
\label{tab:edgetpu_map}
\begin{tabular}{lrc}
\toprule
\textbf{Operator} & \textbf{Count} & \textbf{Status} \\
\midrule
ADD             & 6  & Mapped to Edge-TPU \\
SPLIT           & 4  & Mapped to Edge-TPU \\
MEAN            & 1  & Mapped to Edge-TPU \\
QUANTIZE        & 2  & Mapped to Edge-TPU \\
MUL             & 26 & Mapped to Edge-TPU \\
SOFTMAX         & 1  & Mapped to Edge-TPU \\
CONV\_2D        & 26 & Mapped to Edge-TPU \\
LOGISTIC        & 26 & Mapped to Edge-TPU \\
CONCATENATION   & 4  & Mapped to Edge-TPU \\
PAD             & 5  & Mapped to Edge-TPU \\
FULLY\_CONNECTED& 1  & Mapped to Edge-TPU \\
\bottomrule
\end{tabular}
\end{table}

\paragraph{Reproducibility.}
To support reproducibility, we provide conversion and compilation scripts, environment configuration, and calibration-data
specification in an accompanying implementation repository.\footnote{\url{https://github.com/VesalAhsani/Google-Coral-Edge-TPU-Implementation}} We also report end-to-end runtime measurements in Section~\ref{sec:inv_eval}.

\begin{figure}[t]
  \centering
  \includegraphics[width=\linewidth]{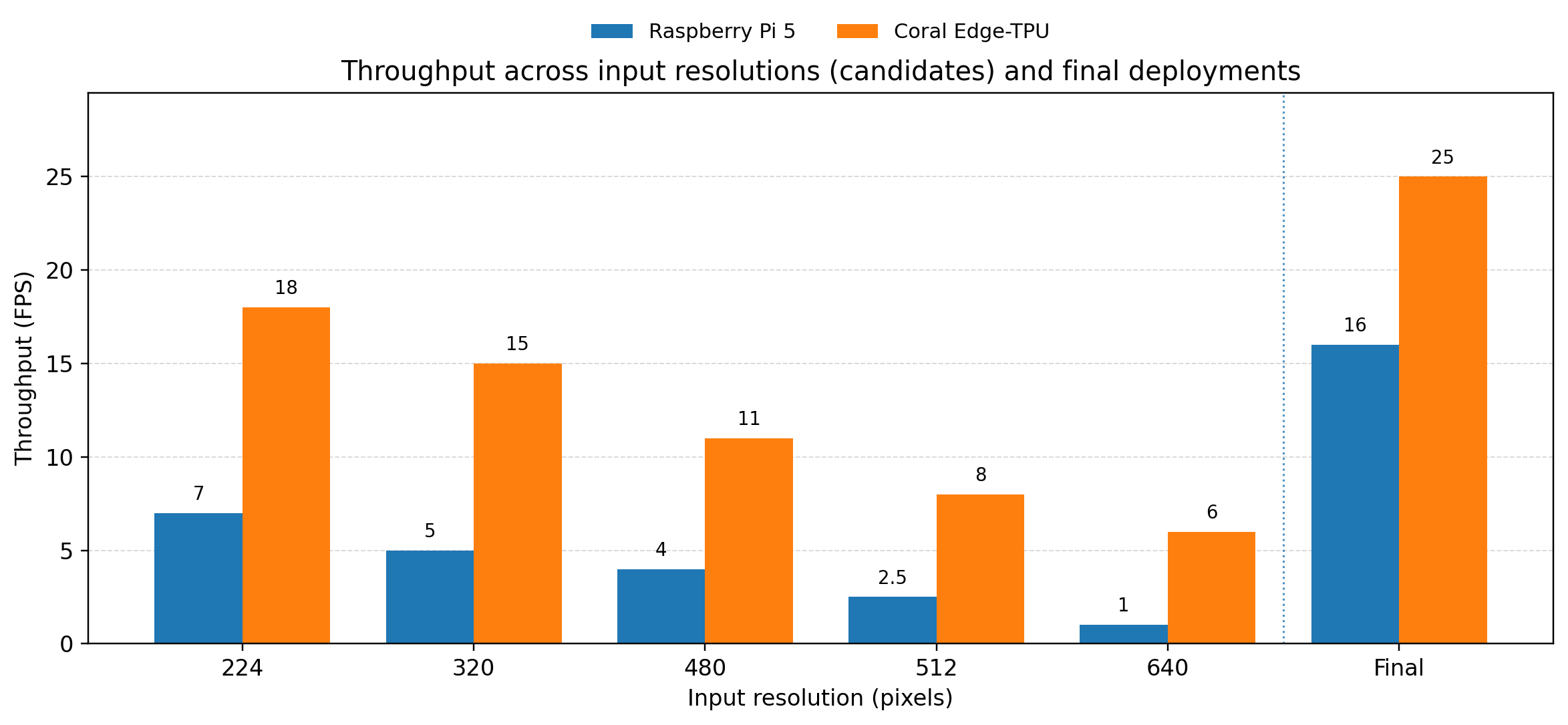}
  \caption{Throughput (FPS) across input resolutions for candidate configurations and final deployments on Raspberry Pi~5 and Coral. Candidate runs sweep the input resolution from 224 to 640 pixels; the dashed separator indicates the final optimized deployment configurations.}
  \label{fig:throughput_resolutions}
\end{figure}

Figure~\ref{fig:throughput_resolutions} reports end-to-end throughput as a function of input resolution on both target platforms. As input resolution increases, throughput decreases on both devices, with a sharper drop on the CPU-only Raspberry Pi~5. The final optimized deployments achieve 16~FPS on Raspberry Pi~5 and 25~FPS on Coral, meeting the throughput requirements for persistence-based alerting.

\subsection{Runtime profiling}
\label{sec:profiling}
To characterize system-level behavior, we profile the pipeline into five stages: capture/decode, preprocessing, inference,
postprocessing (including class mapping and temporal gating), and overlay/I/O. Each stage is instrumented with timestamps
during live operation, and we report mean and representative percentile latencies to capture both typical performance and
tail behavior. Table~\ref{tab:profile} summarizes the runtime breakdown template used for reporting per-stage and end-to-end
latency on each platform.

\begin{table}[h]
\centering
\caption{Runtime breakdown for end-to-end profiling (report mean and percentile latencies).}
\label{tab:profile}
\begin{tabular}{lccc}
\toprule
\textbf{Stage} & \textbf{Pi 5 (ms)} & \textbf{Coral (ms)} & \textbf{Notes} \\
\midrule
Capture + decode & 6  & 6  & camera I/O, optional ROI strategy \\
Preprocess        & 4  & 4  & resize/normalize \\
Inference         & 38 & 22 & INT8 CPU / Edge-TPU \\
Postprocess       & 5  & 5  & mapping + temporal gate \\
Overlay / I/O     & 4  & 3  & UI, logging, event emission \\
\midrule
\textbf{Total}    & 57 & 40 & end-to-end per-frame latency \\
\bottomrule
\end{tabular}
\end{table}

\section{Experiments and Evaluation}
\label{sec:eval}
We evaluate the proposed system at two complementary levels that reflect practical deployment requirements:
(i) frame-level recognition quality of the per-frame classifier, and (ii) event-level alert behavior after temporal
persistence gating (Section~\ref{sec:temporalhead}). This separation is important because a model can achieve strong
frame-level accuracy while still producing an unusable alert stream due to jitter, fragmentation, or frequent nuisance
triggers. Our evaluation therefore emphasizes both recognition performance and the operational properties of the resulting
alerts.

\subsection{Evaluation protocol}
All frame-level recognition results are reported on the held-out test split described in Section~\ref{sec:dataset}.
Event-level evaluation is performed by running the full pipeline (per-frame inference followed by the temporal decision head)
over temporally ordered sequences and comparing produced alert events against reference behavior intervals derived from the frame-level labels (Section~\ref{sec:event_eval}).

\paragraph{Ablation measurement protocol.}
All ablation results in Table~\ref{tab:ablation} are computed on the held-out test split described in Section~\ref{sec:dataset}. Unless otherwise stated, we report the mean over $S{=}3$ independent training runs with different random seeds (same data splits and evaluation protocol). For each variant, we run the identical end-to-end pipeline (capture $\rightarrow$ preprocess $\rightarrow$ per-frame inference
$\rightarrow$ class mapping $\rightarrow$ temporal decision head) over temporally ordered test sequences.
Frame-level metrics (macro-F1) are computed from per-frame predictions $\hat{y}_t=\arg\max_c p_t(c)$.
Event-level metrics are computed from the alert events emitted by the temporal head using a fixed operating point
($\tau{=}0.75$, $K{=}25$ frames, and a fixed cooldown), held constant across variants except when the temporal head is removed
(i.e., $K{=}1$). False-alert rate is reported as unmatched predicted alert events per minute, aggregated over all non-\textit{Normal} classes,
divided by the total evaluated test duration in minutes (Section~\ref{sec:event_eval}).

\subsection{Metrics}
\label{sec:metrics}
\paragraph{Frame-level recognition.}
Let $\hat{y}_t=\arg\max_c p_t(c)$ be the predicted class for frame $t$. We report three complementary metrics:
(i) macro-F1, computed as the unweighted mean of per-class F1 scores to reflect performance under class imbalance,
(ii) balanced accuracy, computed as the mean per-class recall to reduce dominance of the \emph{Normal} class, and
(iii) per-class F1 to identify failure modes and confusions among visually similar behaviors (e.g., phone use versus grooming,
or control-panel interaction versus texting). Macro-F1 and balanced accuracy reflect complementary aspects of recognition quality:
macro-F1 penalizes precision/recall trade-offs for rare classes, while balanced accuracy emphasizes recall consistency across classes.

\begin{figure}[t]
  \centering
  \includegraphics[width=\linewidth]{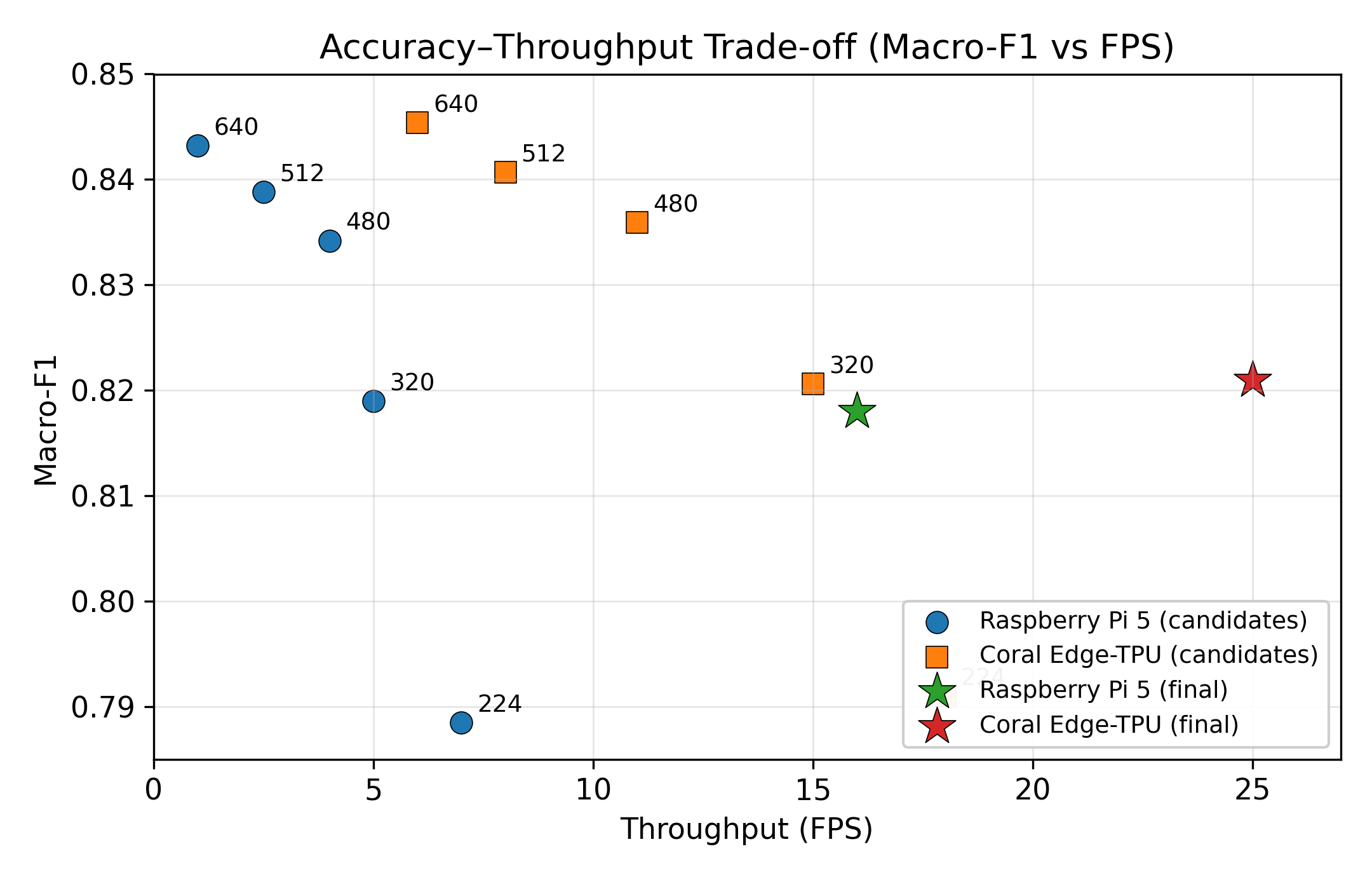}
  \caption{Accuracy--throughput trade-off across candidate input resolutions and final deployments. Each point shows frame-level macro-F1 versus measured throughput (FPS). Candidate points correspond to the resolution sweep on each platform; star markers indicate the final deployed configurations.}
  \label{fig:macroF1_vs_fps}
\end{figure}

Figure~\ref{fig:macroF1_vs_fps} summarizes the accuracy--throughput trade-off across candidate input resolutions on both platforms. Higher resolutions tend to improve macro-F1 only marginally while incurring a substantial throughput penalty, especially on the CPU-only target. The final deployed configurations sit in a regime that preserves competitive macro-F1 while providing real-time throughput, enabling stable event-level alert generation after persistence gating.

\subsubsection{Event-level evaluation and ground-truth (GT) construction}
\label{sec:event_eval}

To obtain reference events for evaluation, we convert the frame-level labels $y_t$ into contiguous intervals by merging
consecutive frames with the same non-\textit{Normal} label. Each maximal contiguous run defines one reference event
$(c, t_\text{start}, t_\text{end})$ for class $c$.

\paragraph{Predicted event stream.}
The deployed system outputs alert events $\mathcal{A}=\{(c,t_\text{start},t_\text{end})\}$ produced by the temporal decision head.

\paragraph{Event matching.}
A predicted event and a GT event of the same class are considered a match if their temporal intersection-over-union (tIoU) exceeds a threshold $\eta$:
$\mathrm{tIoU} = \frac{|I_\text{pred}\cap I_\text{gt}|}{|I_\text{pred}\cup I_\text{gt}|} \ge \eta$.
Unless otherwise stated, we use $\eta{=}0.3$ to tolerate minor boundary offsets introduced by persistence gating.
Each predicted event can match at most one GT event (greedy matching by highest tIoU).

\paragraph{Event metrics.}
False alerts/min is computed as the number of unmatched predicted events aggregated over classes $c\in\{2,\dots,C\}$,
divided by the total evaluated duration in minutes.
Time-to-detect is computed for matched events as $t^\text{pred}_\text{start} - t^\text{gt}_\text{start}$.
Fragmentation is computed as the number of predicted segments matched to a single GT event (lower is better).

\paragraph{Event-level alert quality.}
The temporal decision head produces alert events $\mathcal{A}=\{(c,t_\text{start},t_\text{end})\}$, where each event denotes
a temporally sustained detection of class $c$. We evaluate operational alert behavior using:
(i) false alert rate (alerts per minute), measured on segments without the corresponding ground-truth event,
(ii) mean time-to-detect for sustained events, defined as the delay between the annotated event onset and the alert onset, and
(iii) alert fragmentation (jitter), quantified as the number of emitted alert segments per ground-truth event (lower is better).
Together, these metrics capture the practical trade-off between fast detection and low nuisance alarms and directly reflect
the usability of the alert stream for driver feedback and downstream integration.

\subsection{Baselines}
\label{sec:baselines}
We compare against baselines designed to isolate the contribution of (i) temporal decision logic and (ii) confounder-aware
labeling. Unless otherwise stated, baselines use the same per-frame classifier and the same preprocessing pipeline as the proposed
system, and differ only in how frame-level predictions are converted into alerts. Hyperparameters for temporal smoothing and gating
(e.g., window size, EMA factor, thresholds) are selected on the validation split and then held fixed for test evaluation.

\begin{itemize}[leftmargin=*]
  \item \textbf{Frame-only (no temporal head).}
  This baseline triggers alerts directly from frame-level predictions. Specifically, an alert of class $c$ is emitted at time $t$
  whenever $\hat{y}_t=c$, which is equivalent to $K{=}1$ in Eq.~\ref{eq:gate}. This represents the simplest possible integration of a
  classifier into an online system. In practice, it is highly sensitive to transient motions and short occlusions, and typically produces
  fragmented alerts and nuisance triggers. It serves as a lower bound on operational stability and quantifies the benefit of explicit
  persistence constraints.

  \item \textbf{Temporal smoothing (non-gated).}
  This baseline applies temporal regularization to reduce high-frequency label switching without enforcing a strict persistence requirement.
  We consider two common smoothing strategies: (i) a sliding-window majority vote over recent hard predictions $\hat{y}_{t-w+1:t}$, and
  (ii) an exponential moving average (EMA) over class probabilities,
  $\tilde{\mathbf{p}}_t = \lambda \tilde{\mathbf{p}}_{t-1} + (1-\lambda)\mathbf{p}_t$, followed by per-frame thresholding.
  Smoothing reduces instantaneous jitter, but unlike persistence gating it can still yield alerts during short transient gestures if the
  smoothed score crosses threshold briefly. This baseline isolates the benefit of enforcing sustained evidence (Eq.~\ref{eq:gate}) beyond
  generic temporal smoothing.

  \item \textbf{Landmark-based heuristic (drowsiness proxy).}
  This baseline reflects classical DMS pipelines that rely on face/eye landmarks and heuristic decision rules rather than direct
  multi-behavior image classification. Using MediaPipe \cite{lugaresi2019mediapipe}, we estimate facial landmarks and derive simple
  temporal cues (e.g., sustained eye closure and yawning-related mouth dynamics), then apply threshold-based logic to flag drowsiness.
  This baseline is not intended to cover the full 17-class behavior set; rather, it provides a reference point for robustness under
  occlusion and illumination changes, where landmark tracking can degrade (e.g., hands covering the face, sunglasses, masks, low-light).
\end{itemize}

\subsection{Ablation study}
\label{sec:ablation}
We ablate two design choices: (i) explicit confounder classes in the label taxonomy, and (ii) temporal persistence gating for
alert generation. Each ablation disables one component while keeping the rest of the pipeline unchanged. This isolates how
each mechanism contributes to both recognition quality and operational alert stability.

Table~\ref{tab:ablation} reports the evaluation template. The key expected trend is that confounder-aware labeling primarily
reduces systematic confusions among visually similar actions (improving macro-F1 and reducing certain false alerts), while the
temporal decision head primarily suppresses transient triggers (reducing false alerts/min and alert fragmentation). The full
system combines both benefits by improving recognition reliability and stabilizing alerts under real driving disturbances.

\begin{table}[h]
\centering
\caption{Ablation study variants (report measured results on the held-out test set and event-level evaluation).}
\label{tab:ablation}
\begin{tabular}{lccc}
\toprule
\textbf{Variant} & \textbf{Macro-F1} & \textbf{False alerts/min} & \textbf{Notes} \\
\midrule
No confounders + no temporal head & 0.71 & 1.80 & raw per-frame alerting \\
Confounders only & 0.78 & 1.20 & fewer visual confusions \\
Temporal head only & 0.74 & 0.55 & suppresses transient alarms \\
Confounders + temporal head (full) & 0.82 & 0.30 & proposed system \\
\bottomrule
\end{tabular}
\end{table}

\section{In-Vehicle Evaluation}
\label{sec:inv_eval}
Offline test splits are necessary but not sufficient for deployable in-cabin monitoring: real driving introduces
distribution shifts and system-level effects that are difficult to fully reproduce in curated datasets.
These include rapid illumination transitions (tunnels/shadows, sunset glare), vibration-induced motion blur,
rolling-shutter artifacts, partial occlusions by steering wheel/hands/objects, and small viewpoint changes caused by
mounting geometry. We therefore evaluate the \emph{deployed} system under live in-vehicle operation and report
end-to-end runtime behavior and qualitative alert stability under realistic disturbances.

All results in this section are measured \emph{end-to-end} on-device (camera capture to alert output) and include
the full pipeline described in Section~\ref{sec:system}: capture/decode, preprocessing, per-frame inference,
postprocessing (class mapping + temporal logic), and overlay/event emission. This scope is important because a DMS can
appear fast in inference-only benchmarks but still exhibit unacceptable delay or jitter once camera I/O and application logic are included.\footnote{Supplementary pilot-test videos from live in-vehicle operation:
\url{https://youtu.be/8gRq2xjAEpU}, \url{https://youtu.be/-goZnIqVohw}, \url{https://youtu.be/9cnE6L_SoJY}.}

\subsection{Test setting}
The in-vehicle evaluation uses a single driver-facing in-cabin camera mounted in a dashboard / rear-view-mirror region.
For night and low-light operation, we use the same camera in NIR mode with active IR illumination (integrated IR LEDs),
while daytime operation uses visible-spectrum imaging when available.
Tests are conducted across typical daytime and
mixed-light driving and include natural disturbances such as vehicle motion, brief occlusions, and transient head/hand
movements. These factors correspond to the dominant deployment failure modes for in-cabin vision: short-duration blur,
rapid exposure/contrast shifts, partial occlusion of the face/hands, and moderate viewpoint variation due to mounting.

\begin{figure}[h]
  \centering
  \begin{minipage}[t]{0.49\linewidth}
    \centering
    \includegraphics[width=\linewidth]{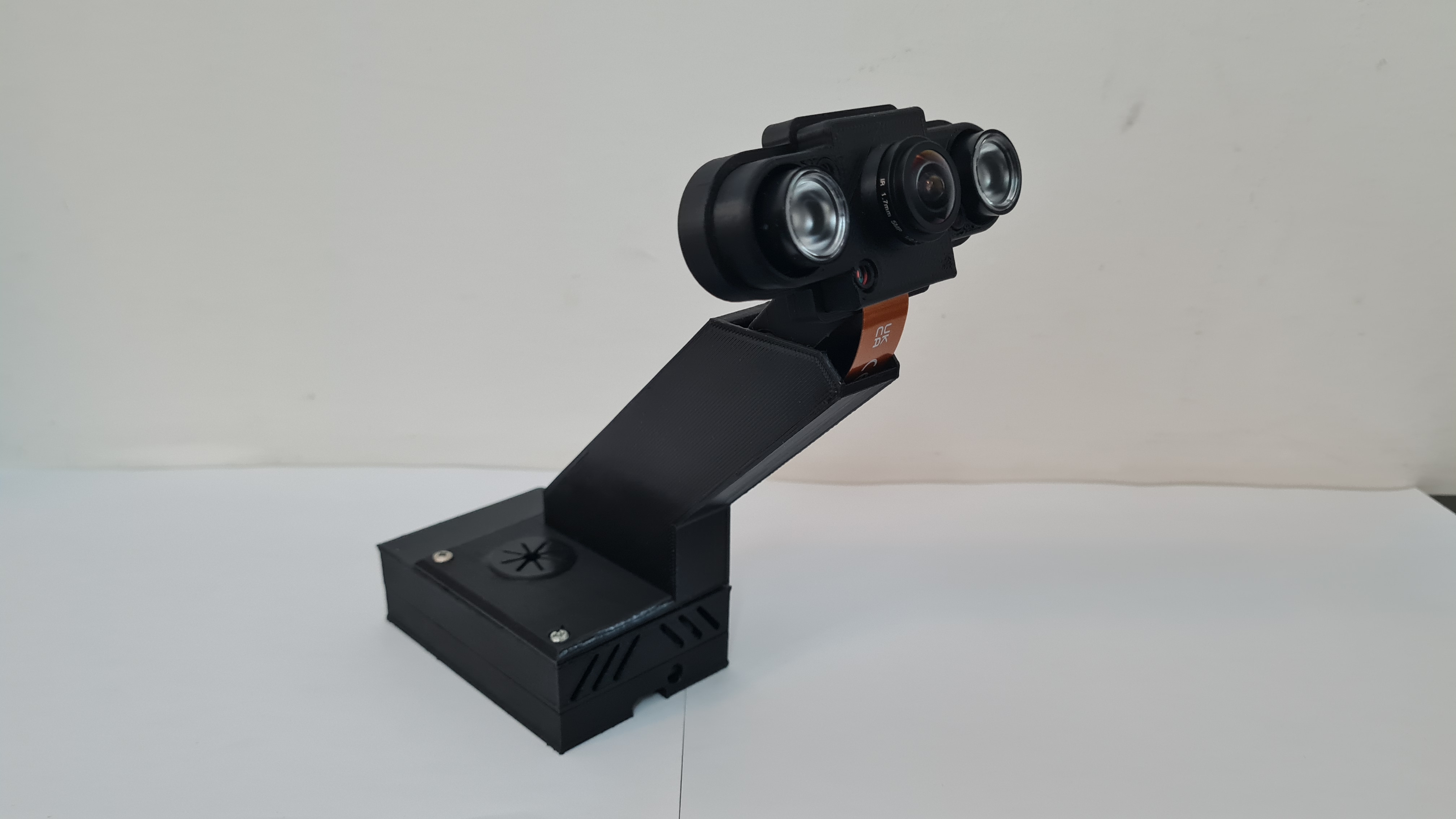}
  \end{minipage}\hfill
  \begin{minipage}[t]{0.49\linewidth}
    \centering
    \includegraphics[width=\linewidth]{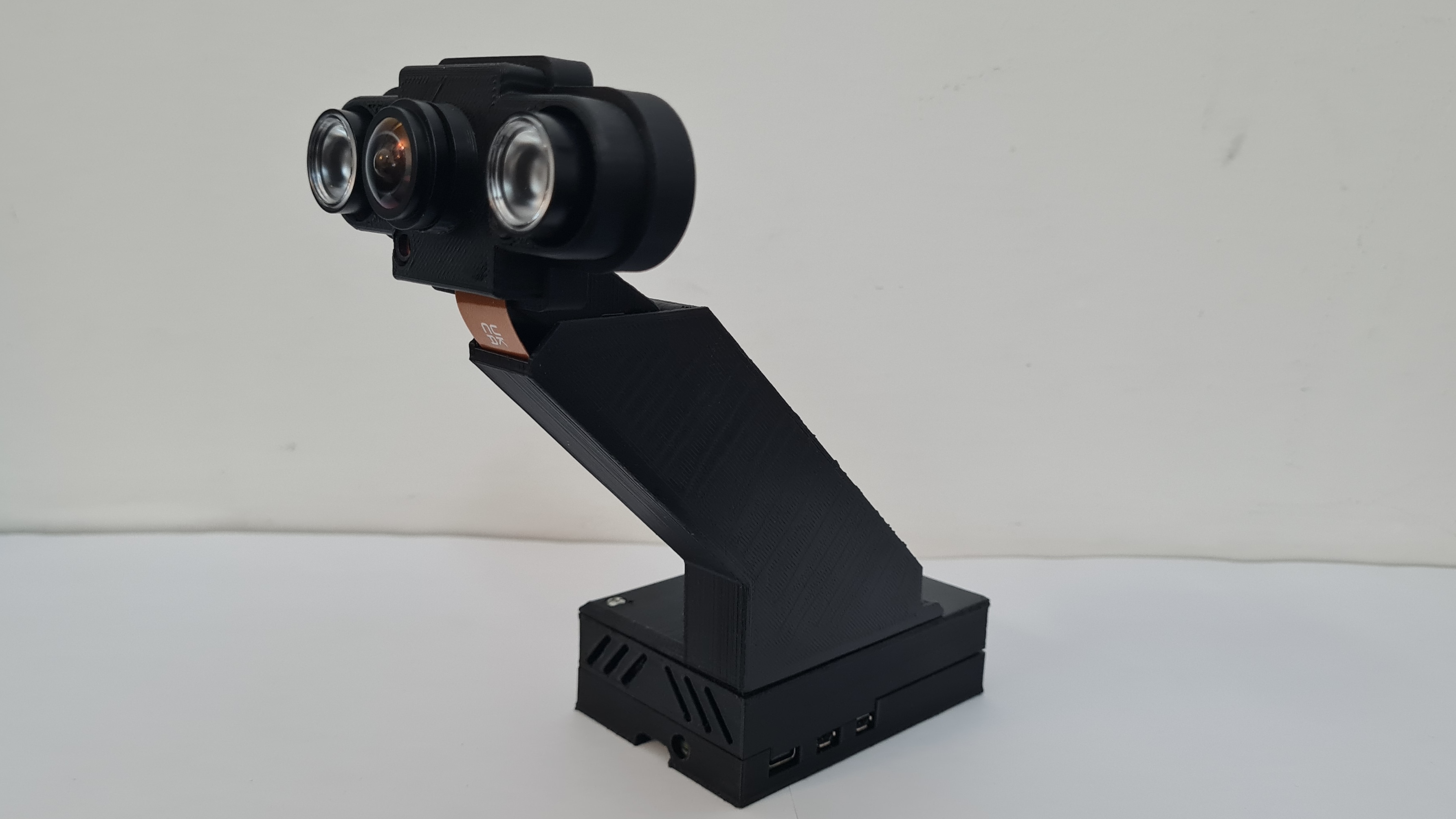}
  \end{minipage}
  \caption{Prototype edge-hardware setups used for live in-vehicle evaluation (camera + compute unit).}
  \label{fig:inv_setups}
\end{figure}

To reflect intended deployment usage, the system runs continuously on the target hardware and processes frames in a
streaming manner. Runtime measurements therefore include capture/decode, preprocessing, inference, postprocessing
(including temporal decision logic), and overlay/event emission. The prototypes used for evaluation are shown in
Figure~\ref{fig:inv_setups}.

\paragraph{Measurement methodology (end-to-end and stage-aware).}
We instrument the runtime loop with per-stage timestamps and compute end-to-end latency as
\begin{equation}
\label{eq:e2e_latency}
L^{\mathrm{e2e}}_t = L^{\text{cap}}_t + L^{\text{pre}}_t + L^{\text{inf}}_t + L^{\text{post}}_t + L^{\text{io}}_t,
\end{equation}
where the terms correspond to capture/decode, preprocessing, inference, postprocessing (including temporal gating),
and overlay/I/O. Throughput (FPS) is measured over continuous streaming operation and reflects sustainable processing
rate rather than short burst timing. This stage-aware measurement aligns with the profiling decomposition reported in
Section~\ref{sec:profiling} and allows us to interpret which components dominate end-to-end behavior on each platform.

\subsection{Runtime results}
Table~\ref{tab:runtime} summarizes end-to-end runtime measured during live operation on the two target platforms.
Compared to an early MVP configuration, optimized deployments achieve near real-time inference and stable alert generation.
On Raspberry Pi~5, the optimized pipeline sustains approximately $\sim$16 FPS with end-to-end latency below 60 ms/frame.
On Coral, after TFLite export and Edge-TPU compilation, the deployed pipeline sustains approximately $\sim$25 FPS with
end-to-end latency of approximately $\sim$40 ms/frame.

\paragraph{Why these numbers matter for alerting.}
The temporal decision head (Section~\ref{sec:temporalhead}) operates in \emph{frames}: it requires sustained evidence over a
window of length $K$ frames above confidence threshold $\tau$.
Because $K$ is measured in frames, the same $K$ corresponds to different \emph{time} windows depending on realized FPS.
With the default setting $K{=}25$, the effective persistence window is approximately:
\begin{itemize}[leftmargin=*]
  \item Raspberry Pi~5: $25/16 \approx 1.56$ seconds,
  \item Coral Edge-TPU: $25/25 = 1.00$ seconds.
\end{itemize}
Thus, the reported FPS is not only a throughput statistic: it directly determines the real-time behavior of event triggering.
In practice, if a fixed time window is desired across platforms, $K$ can be scaled by the measured FPS while keeping the same
operational semantics.

\paragraph{Latency vs.\ throughput trade-off at the system level.}
The end-to-end latency values in Table~\ref{tab:runtime} represent the time between receiving a frame and producing an updated
alert state (and optional overlay). For sustained behaviors, the dominant contributor to time-to-alert is typically the persistence
requirement (the $K$-frame window), while $L_{\mathrm{e2e}}$ contributes an additional tens of milliseconds per update. This separation
is intentional: the temporal head trades immediate reaction to single-frame spikes for stable event-level alerts.

\begin{table}[h]
\centering
\caption{End-to-end runtime performance measured during live in-vehicle operation.}
\label{tab:runtime}
\begin{tabular}{lcc}
\toprule
\textbf{Deployment} & \textbf{End-to-end latency (ms/frame)} & \textbf{Throughput (FPS)} \\
\midrule
Early MVP (best observed) & $\sim$125--200 & $\sim$5--8 \\
Raspberry Pi~5 (INT8) & $<60$ & $\sim$16 \\
Coral Edge-TPU (INT8, compiled) & $\sim$40 & $\sim$25 \\
\bottomrule
\end{tabular}
\end{table}

\paragraph{Interpreting the improvement from MVP to optimized deployment.}
The early MVP regime (5--8 FPS) is insufficient for reliable persistence-based alerting because temporal gating requires a
consistent frame cadence to represent short windows faithfully and to avoid unstable event boundaries. Moving to $\sim$16 FPS (Pi~5)
and $\sim$25 FPS (Coral) increases the temporal resolution of the alert logic and reduces the probability that brief motion artifacts
dominate the alert stream. This is also consistent with the profiling view in Section~\ref{sec:profiling}, where inference is the primary
bottleneck on the CPU-only target, while accelerator execution shifts the operating point toward lower per-frame latency and higher sustained throughput.

\subsection{Qualitative robustness and operational stability}
Beyond throughput, we qualitatively assessed alert stability during extended driving segments, focusing on failure modes that commonly
degrade usability in practice: transient motions, short glance shifts, brief hand movements, and occlusions from objects such as phones,
cups, and the steering wheel. The qualitative analysis focuses on \emph{event behavior} (stable segments with meaningful onset/offset)
rather than per-frame labels, since real deployments consume alerts as events.

\begin{figure}[H]
  \centering
  \includegraphics[width=\linewidth]{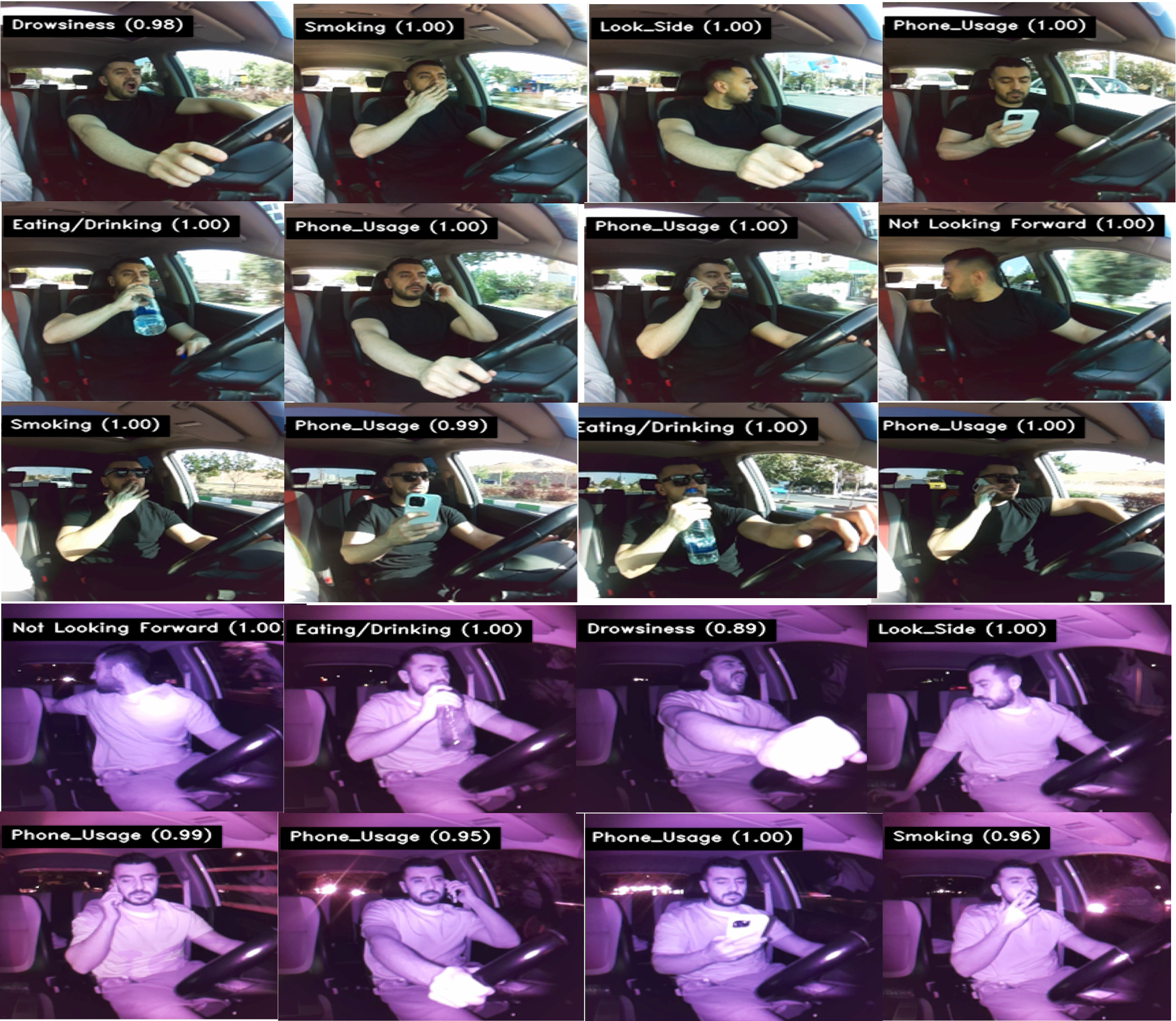}
  \caption{Qualitative snapshots from live in-vehicle deployment. Each panel shows the on-device overlay with the top-1 predicted behavior label and its associated confidence score (softmax probability). The examples span daytime RGB driving and night operation using the NIR camera with active IR illumination (purple tone), and include representative distraction/drowsiness-related behaviors (e.g., phone use, eating/drinking, smoking, off-road gaze, and drowsiness cues). These snapshots illustrate that the deployed pipeline produces confident per-frame predictions under practical disturbances such as illumination change, glare/low-light, accessories (e.g., sunglasses), and partial occlusions, providing stable evidence for the event-level alert logic.}
  \label{fig:inv_examples}
\end{figure}

Figure~\ref{fig:inv_examples} shows representative frames from live in-vehicle runs with the real-time prediction overlay.
We include both daytime RGB and night/low-light operation, capturing common deployment disturbances such as rapid exposure
change, vibration-induced blur, and partial occlusion. The displayed labels correspond to the application-level behavior
outputs used for monitoring/alerting in the deployed prototype. In this specific in-vehicle visualization, we apply an
optional class-mapping layer that groups the underlying 17-class predictions into a smaller set of application categories
(eating/drinking, not looking forward, smoking, look-side, drowsiness, and phone usage) to simplify downstream alerting and
user-facing display. Importantly, the per-frame model still predicts all 17 behaviors distinctly; the mapping is a purely
postprocessing choice and can be customized to (i) preserve all 17 classes, or (ii) merge selected classes depending on the
target alert policy and deployment requirements.\footnote{Supplementary in-vehicle pilot-test videos recorded during live operation are available at: \url{https://youtu.be/8gRq2xjAEpU}, \url{https://youtu.be/-goZnIqVohw}, and \url{https://youtu.be/9cnE6L_SoJY}.}

\paragraph{Effect of persistence gating on nuisance triggers.}
Without persistence constraints (equivalently $K{=}1$), brief motions and short occlusions can produce isolated misclassifications that
fragment alerts and inflate nuisance alarms. With confidence-and-persistence gating enabled, these transient spikes are suppressed unless
the model remains confident over a sustained window. In live operation, this produces fewer single-frame ``pops'' and yields alert segments
that better align with the intended semantics of sustained distraction/drowsiness behaviors rather than incidental gestures.

\paragraph{Confounder-aware taxonomy reduces systematic false positives.}
The confounder-aware label design reduces a common deployment failure: benign but visually similar behaviors being absorbed by nearby
safety-critical categories. Explicit classes such as grooming/hand-on-hair and control-panel interaction provide alternative explanations
for hand-to-face and hand-to-console postures that otherwise resemble phone use or texting. In live operation, this reduces spurious phone-related
triggers during benign interaction patterns and improves the usability of the alert stream, especially when small objects are partially occluded.

\paragraph{Operational implications: stable alerts under real disturbances.}
The combined system (confounder-aware recognition + temporal head) is designed to tolerate the two major sources of instability in real driving:
(i) visual intermittency (occlusion, blur, illumination flicker) and (ii) behavior intermittency (natural brief glances/gestures).
While this evaluation is primarily intended to validate runtime behavior and qualitative stability rather than to provide a full on-road
benchmark, the observed end-to-end performance and alert behavior support the central deployment claim of this work: stable, real-time
in-cabin behavior recognition and persistence-gated alert generation on low-cost embedded hardware under realistic in-vehicle disturbances.

\section{Conclusion}
\label{sec:conclusion}
We presented an end-to-end, single-camera in-cabin driver behavior recognition system designed explicitly for
real-time deployment on low-cost edge hardware. Rather than optimizing only offline frame-level accuracy, the
proposed design treats deployability as a system constraint and targets stable, actionable alerts under real
in-cabin variability. The pipeline combines (i) a compact per-frame vision model, (ii) a confounder-aware 17-class
behavior taxonomy that reduces visually induced false positives among look-alike actions, and (iii) a temporal
decision head that converts noisy per-frame probabilities into event-level alerts via confidence-and-persistence
gating.

We validated the full on-device pipeline under live in-vehicle operation on two embedded targets: a CPU-only
Raspberry Pi~5 and a Google Coral development board with an Edge~TPU accelerator. End-to-end measurements
(including capture, preprocessing, inference, postprocessing, and output) show near real-time performance on both
platforms, achieving approximately $\sim$16 FPS on Raspberry Pi~5 with INT8 inference (per-frame latency below
60~ms) and approximately $\sim$25 FPS on Coral with compiled INT8 execution (end-to-end latency $\sim$40~ms).
These operating points provide sufficient temporal resolution for persistence-based alerting and demonstrate that
fine-grained in-cabin behavior recognition can be deployed on inexpensive hardware without relying on cloud
inference.

Overall, the results support the central claim of this work: with confounder-aware labeling and persistence-gated
alerting, a compact vision pipeline can produce a stable, usable behavior-alert stream under realistic driving
disturbances (illumination shifts, vibration, occlusion) while meeting tight latency and throughput constraints
on low-cost edge platforms. This establishes a practical foundation for continuous in-cabin monitoring in
cost-constrained vehicle pilots and for downstream integration scenarios that require temporally consistent,
event-level behavioral evidence rather than jittery frame-by-frame labels.

\section{Future Work and Outlook (AgV)}
\label{sec:future}
This paper establishes a deployable driver monitoring system (DMS) that produces temporally stable, event-level
behavior signals on low-cost edge hardware. A natural and research-driven next step is to generalize this driver-only
capability toward an \emph{occupant monitoring system} (OMS) that models the full in-cabin human context and provides a
richer upstream signal for higher-level vehicle intelligence, including emerging agentic vehicle (AgV) directions
\cite{yu2025agv}.

\paragraph{From driver behaviors to cabin-level human context (DMS $\rightarrow$ OMS).}
Whereas DMS focuses on a single subject (the driver) and a limited set of safety-relevant behaviors, OMS requires
reasoning over multiple occupants, their states, and their interactions. The technical shift is not merely increasing
the number of classes; it requires representing \emph{who} is doing \emph{what}, \emph{where}, and \emph{for how long} in
a shared, occlusion-heavy space. Research questions include: (i) efficient multi-occupant perception under a single
camera viewpoint, (ii) robust separation of concurrent activities, and (iii) stable temporal aggregation into a compact
cabin-state representation suitable for downstream logic.

\paragraph{Cabin-state representation and temporal semantics.}
To act as an input to vehicle intelligence, OMS outputs should move beyond frame-level labels and even beyond isolated
driver alerts. A promising direction is a structured cabin-state representation that captures: occupant presence and
identity (driver vs.\ passengers), per-occupant activity states (e.g., distraction, interaction, posture), and temporal
properties (persistence, transitions, and event boundaries). Building on the persistence-gated design in this work,
future research can formalize cabin-state estimation as a temporal inference problem with explicit semantics for onset,
offset, and stability---so that downstream modules consume consistent evidence rather than jittery per-frame predictions.

\paragraph{Toward human-centered vehicle intelligence and AgV integration.}
AgV perspectives emphasize decision-making that adapts to human context rather than relying only on fixed
automation rules \cite{yu2025agv}. In this framing, OMS provides the upstream evidence needed to condition higher-level
policies: estimating driver readiness and engagement, timing interactions appropriately, and modulating assistance when
occupant context indicates elevated risk or reduced attention. Concretely, OMS-derived cabin state can support hierarchical
policies that differentiate transient, benign events from sustained secondary-task engagement and use this distinction to
trigger graded responses (e.g., defer non-critical prompts, escalate warnings, or adjust assistance timing). At the same time,
cross-layer analyses of AgVs highlight that higher-level decision layers can become sensitive to systematic errors or
misinterpretations in upstream perception signals, motivating robustness mechanisms before such integration
\cite{eslami2025agvsec}. This positions
OMS not as a standalone alerting module, but as a perception layer that supplies structured, temporally reliable human-context
signals for decision layers.

\paragraph{Research trajectory.}
In summary, our planned research direction is to evolve the presented edge-deployable DMS into an OMS that estimates a
structured, temporally stable cabin-state representation and to study how such representations improve higher-level vehicle
intelligence. This trajectory connects low-level in-cabin perception to human-centered decision-making, providing a concrete
path from deployable sensing to the broader goal of context-aware, agentic mobility systems.


\begin{thebibliography}{99}

\bibitem{cdc_drowsy_2013}
Centers for Disease Control and Prevention (CDC).
\newblock Drowsy driving and risk behaviors --- 10 states and Puerto Rico, 2011--2012.
\newblock \emph{MMWR Morbidity and Mortality Weekly Report}, 63(26):557--562, 2014.

\bibitem{nhtsa_drowsy_2021}
National Highway Traffic Safety Administration (NHTSA).
\newblock Drowsy driving.
\newblock Web page.
\newblock \url{https://www.nhtsa.gov/risky-driving/drowsy-driving} (accessed 2026-01-06).

\bibitem{nhtsa_distracted_2021}
National Highway Traffic Safety Administration (NHTSA).
\newblock Distracted driving.
\newblock Web page.
\newblock \url{https://www.nhtsa.gov/risky-driving/distracted-driving} (accessed 2026-01-06).

\bibitem{sae_j3016}
SAE International.
\newblock Taxonomy and definitions for terms related to driving automation systems for on-road motor vehicles (SAE~J3016).
\newblock Standard, revised 2021.
\newblock \url{https://www.sae.org/standards/content/j3016_202104/} (accessed 2026-01-06).

\bibitem{eu_2019_2144}
European Parliament and Council of the European Union.
\newblock Regulation (EU) 2019/2144 of 27 November 2019 on type-approval requirements for motor vehicles and their trailers, and systems, components and separate technical units intended for such vehicles (General Safety Regulation).
\newblock \emph{Official Journal of the European Union}, L~325, 2019.
\newblock \url{https://eur-lex.europa.eu/eli/reg/2019/2144/oj} (accessed 2026-01-06).

\bibitem{eu_2021_1341}
European Commission.
\newblock Commission Delegated Regulation (EU) 2021/1341 of 23 April 2021 supplementing Regulation (EU) 2019/2144 by laying down detailed rules concerning specific test procedures and technical requirements for type-approval of motor vehicles with regard to their emergency lane-keeping systems, advanced emergency braking systems, driver drowsiness and attention warning systems, etc.
\newblock 2021.
\newblock \url{https://eur-lex.europa.eu/eli/reg_del/2021/1341/oj} (accessed 2026-01-06).

\bibitem{unece_ddaw}
United Nations Economic Commission for Europe (UNECE), WP.29 Vehicle Regulations.
\newblock Driver Drowsiness and Attention Warning (DDAW).
\newblock Web page.
\newblock \url{https://unece.org/transport/vehicle-regulations/driver-drowsiness-and-attention-warning-ddaw} (accessed 2026-01-06).

\bibitem{euroncap_safe_driving}
Euro NCAP.
\newblock Safe driving (ratings/protocol hub).
\newblock Web page.
\newblock \url{https://www.euroncap.com/en/vehicle-safety/the-ratings-explained/safe-driving/} (accessed 2026-01-06).

\bibitem{euroncap_driverengagement}
Euro NCAP.
\newblock Safe driving: Driver engagement.
\newblock Web page.
\newblock \url{https://www.euroncap.com/en/vehicle-safety/the-ratings-explained/safe-driving/driver-engagement/} (accessed 2026-01-06).

\bibitem{euroncap_sd202}
Euro NCAP.
\newblock SD-201: Driver monitoring dossier guidance (technical bulletin), Version~1.0.
\newblock 2025.
\newblock \url{https://www.euroncap.com/media/85788/sd-201-driver-monitoring-dossier-guidance-v10.pdf} (accessed 2026-01-06).

\bibitem{aaa_adas_report}
AAA Foundation for Traffic Safety.
\newblock Vehicle owners' experiences with and perceptions of advanced driver assistance systems.
\newblock Technical report, 2018.
\newblock \url{https://aaafoundation.org/vehicle-owners-experiences-perceptions-advanced-driver-assistance-systems/} (accessed 2026-01-06).

\bibitem{yu2025agv}
J.~Yu.
\newblock Agentic vehicles for human-centered mobility.
\newblock \emph{arXiv preprint arXiv:2507.04996}, 2025.
\newblock doi: 10.48550/arXiv.2507.04996.

\bibitem{eslami2025agvsec}
A.~Eslami and J.~Yu.
\newblock Security risks of agentic vehicles: A systematic analysis of cognitive and cross-layer threats.
\newblock \emph{arXiv preprint arXiv:2512.17041}, 2025.
\newblock doi: 10.48550/arXiv.2512.17041.

\bibitem{donmez2006attitudes}
B.~Donmez, L.~N.~Boyle, J.~D.~Lee, and D.~V.~McGehee.
\newblock Drivers' attitudes toward imperfect distraction mitigation strategies.
\newblock \emph{Transportation Research Part F: Traffic Psychology and Behaviour}, 9(6):387--398, 2006.
\newblock doi: 10.1016/j.trf.2006.02.001.

\bibitem{forster2024_attentionwarn}
M.~F{\"o}rster, N.~Sch{\"o}mig, and M.~Kremer.
\newblock Attentional warnings caused by driver monitoring systems: How often do they appear and how well are they understood?
\newblock \emph{Accident Analysis \& Prevention}, 205:107684, 2024.
\newblock doi: 10.1016/j.aap.2024.107684.

\bibitem{lugaresi2019mediapipe}
C.~Lugaresi et~al.
\newblock MediaPipe: A framework for building perception pipelines.
\newblock \emph{arXiv preprint arXiv:1906.08172}, 2019.
\newblock doi: 10.48550/arXiv.1906.08172.

\bibitem{dinges1998perclos}
D.~F.~Dinges and R.~Grace.
\newblock PERCLOS: A valid psychophysiological measure of alertness as assessed by psychomotor vigilance.
\newblock Technical report, U.S. Federal Highway Administration (FHWA), 1998.

\bibitem{soukupova2016ear}
T.~Soukupov{\'a} and J.~{\v{C}}ech.
\newblock Real-time eye blink detection using facial landmarks.
\newblock In \emph{Computer Vision Winter Workshop (CVWW)}, 2016.

\bibitem{abtahi2014yawdd}
S.~Abtahi, M.~Omidyeganeh, A.~Shirmohammadi, and M.~Hariri.
\newblock YawDD: A yawning detection dataset.
\newblock In \emph{ACM Multimedia Systems Conference (MMSys)}, 2014.

\bibitem{nthu_ddd}
C.-H.~Weng, Y.-H.~Lai, and S.-H.~Lai.
\newblock Driver drowsiness detection via a hierarchical temporal deep belief network.
\newblock In \emph{Asian Conference on Computer Vision (ACCV)}, 2016.

\bibitem{martin2019driveact}
M.~Martin et~al.
\newblock Drive\&Act: A multi-modal dataset for fine-grained driver behavior recognition in autonomous vehicles.
\newblock In \emph{Proceedings of the IEEE/CVF International Conference on Computer Vision (ICCV)}, 2019.

\bibitem{ortega2020dmd}
J.~D.~Ortega et~al.
\newblock DMD: A large-scale multi-modal driver monitoring dataset for attention and alertness analysis.
\newblock In \emph{European Conference on Computer Vision Workshops (ECCVW)}, 2020.
\newblock doi: 10.1007/978-3-030-66823-5\_23.

\bibitem{abouelnaga2017distracted}
Y.~Abouelnaga, H.~M.~Eraqi, and M.~N.~Moustafa.
\newblock Real-time distracted driver posture classification.
\newblock \emph{arXiv preprint arXiv:1706.09498}, 2017.
\newblock doi: 10.48550/arXiv.1706.09498.

\bibitem{wang2023hundreddriver}
J.~Wang, W.~Li, F.~Li, et~al.
\newblock 100-Driver: A large-scale, diverse dataset for distracted driver classification.
\newblock \emph{IEEE Transactions on Intelligent Transportation Systems}, 2023.
\newblock doi: 10.1109/TITS.2023.3255923.

\bibitem{lyu2018longterm}
J.~Lyu, Y.~Li, and M.~Huang.
\newblock Long-term multi-granularity deep framework for driver drowsiness detection.
\newblock \emph{arXiv preprint arXiv:1801.02325}, 2018.
\newblock doi: 10.48550/arXiv.1801.02325.

\bibitem{yu2019conditionadaptive}
J.~Yu, S.~Park, S.~Lee, and M.~Jeon.
\newblock Driver drowsiness detection using a condition-adaptive representation learning framework.
\newblock \emph{arXiv preprint arXiv:1910.09722}, 2019.
\newblock doi: 10.48550/arXiv.1910.09722.

\bibitem{shen2020drowsy}
Q.~Shen et~al.
\newblock Robust two-stream multi-feature network for driver drowsiness detection.
\newblock \emph{arXiv preprint arXiv:2010.06235}, 2020.
\newblock doi: 10.48550/arXiv.2010.06235.

\bibitem{howard2019mobilenetv3}
A.~Howard et~al.
\newblock Searching for MobileNetV3.
\newblock \emph{arXiv preprint arXiv:1905.02244}, 2019.
\newblock doi: 10.48550/arXiv.1905.02244.

\bibitem{tan2019efficientnet}
M.~Tan and Q.~V.~Le.
\newblock EfficientNet: Rethinking model scaling for convolutional neural networks.
\newblock In \emph{Proceedings of the International Conference on Machine Learning (ICML)}, 2019.
\newblock doi: 10.48550/arXiv.1905.11946.

\bibitem{jacob2018quantization}
B.~Jacob et~al.
\newblock Quantization and training of neural networks for efficient integer-arithmetic-only inference.
\newblock \emph{arXiv preprint arXiv:1712.05877}, 2017.
\newblock doi: 10.48550/arXiv.1712.05877.

\bibitem{tflite}
TensorFlow.
\newblock TensorFlow Lite.
\newblock Web page.
\newblock \url{https://www.tensorflow.org/lite} (accessed 2026-01-06).

\bibitem{coralbenchmarks}
Google Coral.
\newblock Edge TPU performance benchmarks.
\newblock Web page.
\newblock \url{https://coral.ai/docs/edgetpu/benchmarks/} (accessed 2026-01-06).

\bibitem{hariharan2023edgedms}
J.~Hariharan, R.~R.~Varior, and S.~Karunakaran.
\newblock Real-time driver monitoring systems on edge AI device.
\newblock \emph{arXiv preprint arXiv:2304.01555}, 2023.
\newblock doi: 10.48550/arXiv.2304.01555.

\bibitem{khalil2025lowcost}
H.~A.~Khalil, O.~M.~Hammad, H.~Abd~El~Munim, and H.~A.~Maged.
\newblock Low-cost driver monitoring system using deep learning.
\newblock \emph{IEEE Access}, 13:14151--14164, 2025.
\newblock doi: 10.1109/ACCESS.2025.3530296.

\bibitem{iso_21448}
ISO.
\newblock ISO~21448:2022 --- Road vehicles --- Safety of the intended functionality (SOTIF).
\newblock Standard, 2022.
\newblock \url{https://www.iso.org/standard/77490.html} (accessed 2026-01-06).

\end{thebibliography}
\end{document}